\pdfoutput=1

\documentclass[11pt]{article}

\usepackage{naacl2021}

\usepackage{times}
\usepackage{latexsym}

\usepackage[T1]{fontenc}

\usepackage[utf8]{inputenc}

\usepackage{booktabs}
\usepackage{siunitx}
\usepackage{algorithm}
\usepackage{algorithmic}
\usepackage{tabularx}
\usepackage{mathtools, nccmath}

\usepackage{amssymb}
\usepackage{amsmath}
\usepackage{bbold}

\usepackage{multirow}
\usepackage{CJKutf8}
\usepackage{microtype}
\usepackage{enumitem}
\usepackage{bm}

\newcommand{\bertbaseqa}{{\sc BERTb$_{QA}$}}
\newcommand{\bertbase}{{\sc BERTb}}
\newcommand{\bertqa}{{\sc mBERT$_{QA}$}}

\newcommand{\mtbertqa}{{\sc mt-mBERT$_{QA}$}}

\newcommand{\mbert}{{\sc mBERT}}
\newcommand{\eg}{\textit{e.g.}}

\definecolor{deepblue}{rgb}{0.16, 0.32, 0.75}
\definecolor{indiagreen}{rgb}{0.00, 0.44, 0.00}
\usepackage{todonotes}

\title{Are Multilingual BERT models robust? A Case Study on Adversarial Attacks for Multilingual Question Answering}

\author{
    Sara Rosenthal,
    Mihaela Bornea,
    Avirup Sil
    \\
    IBM Research AI\\
    Thomas J. Watson Research Center, Yorktown Heights, NY 10598\\
    \{sjrosenthal, mabornea, avi\}@us.ibm.com
   
}

\date{}

\begin{document}
\maketitle
\begin{abstract}
Recent approaches have exploited weaknesses in monolingual question answering (QA) models by adding adversarial statements to the passage. These  attacks  caused  a  reduction  in  state-of-the-art  performance  by  almost  50\%. In this paper, we are the first to explore and successfully attack a multilingual QA (MLQA) system pre-trained on multilingual BERT using several attack strategies for the adversarial statement reducing performance by as much as 85\%. We show that the model gives priority to English and the language of the question regardless of the other languages in the QA pair. Further, we also show that adding our attack strategies during training helps alleviate the attacks.
\end{abstract}

\section{Introduction}
\label{sec:intro}
Most recent advances in question answering (QA) have focused on achieving state-of-the-art results on English datasets \cite{Rajpurkar_2016,rajpurkar2018know,yang2018hotpotqa,Kwiatkowski2019NaturalQA}. 
In this work, we focus on multilingual QA using the MLQA dataset \cite{lewis2019mlqa} which contains parallel examples in seven languages. The MLQA dataset only contains a dev and test set, thus they apply a zero-shot (ZS) approach by training using SQuAD v1.1 \cite{Rajpurkar_2016}, which is in English, and achieve impressive results. 

 However, \cite{jia2017adversarial, wang2018robust} showed that SQuAD models \cite{hu2017reinforced} are fragile when presented with adversarially generated data. 
They proposed AddSentDiverse, which produces a semantically meaningless sentence containing a \textit{fake} answer that looks like the question grammatically, and adds it to the context. Most recently ~\cite{Maharana2020AdversarialAP} has extended the attacks shown in ~\cite{wang2018robust} and has shown a comprehensive analysis focused on English QA, but attacks on multilingual QA has still not been explored. 

\noindent We ask the following research questions:

 \noindent \textbf{(1) Are multilingual QA models trained with \mbert{} robust?} Our main focus is exploring adversarial attacks in a multlingual setting. Therefore, we specifically target \mbert{}. We show that all of our approaches successfully attack the \mbert{} model in all the MLQA languages. Further, our adversarial attacks have a stronger impact on \mbert{} compared to BERT. \\
    
 
 \noindent  \textbf{(2) Does \mbert{} give preference to certain languages?} Although the expectation is that a multilingual language model should treat all languages the same, we hypothesize that the model may prefer to find an answer in certain languages. This could be based on the size of the data used to build the embeddings in \mbert{}, the language of the training data used to fine-tune \mbert{}, or the language of the question in the example. We empirically show the effect of our attacks for all language combinations in MLQA to address this. Our experiments show that when the adversarial statement is in the language of the question the attack has the largest impact. In addition, \mbert{} gives priority to English regardless of the language the question is asked in. We find similar trends when augmenting our \mbert{} model with translated data during training though the MT data does alleviate the priority given to English in some cases.
    
 \noindent  \textbf{(3) Can training with adversarial data alleviate some of the susceptibility to attacks?} After  we have shown that we can successfully attack the MLQA system we turn to addressing the flaws in the model by teaching the model to overcome the attacks. We show that including adversarial statements during training helps alleviate the attacks, especially when using data augmentation strategies to translate our adversarial MLQA data into other languages. 



\begin{figure}[t]
    \centering
    \small
\framebox{%
  \begin{minipage}{\columnwidth}
    \textbf{Wikipedia Page:} JEdit
\begin{center}
    \textbf{English}
\end{center}
\textbf{Context:} \textcolor{red}{\textit{Homeostasis is an example of a programming language used to write Aeronautics.}} The application is highly customizable and can be extended with macros written in BeanShell, Jython, JavaScript and some other scripting languages.
 \\
\textbf{Question:} What is an example of a programming language used to write macros?\\
\textbf{Correct prediction}: \textcolor{indiagreen}{\textbf{BeanShell, Jython, JavaScript}} \\
\textbf{Attack prediction}: \textcolor{red}{\textit{Homeostasis}}
\begin{center}
    \textbf{multilingual $Q_{de}$, $C_{es}$, $S_{en}$}
\end{center}
\textbf{Context:} jEdit se puede personalizar y extender con macros escritas en BeanShell, Jython, JavaScript y otros lenguajes script. \textcolor{red}{\textit{RCA Records is an example of a TBD used to write macros.}}
 \\
\textbf{Question:} Was ist ein Beispiel für eine Programmiersprache, mit der Makros geschrieben werden? \\
\textbf{Correct prediction}: \textcolor{indiagreen}{\textbf{BeanShell, Jython, JavaScript}} \\
\textbf{Attack prediction}: \textcolor{red}{\textit{RCA Records}}
  \end{minipage}}    
  \caption{Examples of attacks on the MLQA dataset. We show two successful attacks on parallel examples, one in English and one with the Question, Context, and \textcolor{red}{\textit{Adversarial Statement}} all in different languages.}
    \label{fig:mlqa-examples}
\end{figure}

We explore these questions by showing that adversarial attacks affect the MLQA dataset by proposing four multilingual adversaries that build on top of prior work \cite{jia2017adversarial, wang2018robust, Maharana2020AdversarialAP} 
and show that it is possible to attack 7 languages in a ZS multi-lingual QA task. We bring down the average performance by at least 20.3\% to as much as 85.6\% depending on the attack used. 
Some examples of attacks are shown in Figure~\ref{fig:mlqa-examples}. 
Finally, after attacking the system we also explore defense strategies to build a model that is less susceptible to attacks. To the best of our knowledge, no recent work has explored adversarial evaluation in a QA setting built on top of large multilingual pre-trained LMs \eg{} \mbert{} \cite{Devlin2018BERTPO}. \\

%

\noindent Our main contribution in this paper is exposing flaws in multilingual QA systems and providing insights that are not evident in a single language:
\begin{itemize}[leftmargin=*]
\item \mbert{} is more susceptible to attacks compared to BERT
\item \mbert{} gives priority to finding the answer in certain languages causing successful attacks even when the adversarial statement is in a different language than the question and context. 
\item Further, \mbert{} gives priority to the language of the question over the language of the context.
\item Augmenting the system with machine translated data helps build a more robust system. 
\end{itemize}
\section{Related Work}
\label{sec:related-work}



There are several prior strategies \cite{jia2017adversarial,wang2018robust,wallace2019universal,Maharana2020AdversarialAP} that introduce adversarial sentences to distract RC systems \cite{hu2017reinforced,shao2018drcd}) reducing SOTA performance by almost 50\%. 
\cite{sen-saffari-2020-models} don't add adversarial statements but analyze how BERT QA models handle missing or incorrect data and question variations. However, all this work focuses on English attacks. Further, although many multilingual QA datasets exist \cite{he2017dureader,asai2018multilingual,mozannar2019neural,artetxe2019cross,lewis2019mlqa}, no prior work has explored adversarial evaluation and exposed vulnerabilities over large pre-trained multi-lingual language models. \cite{Devlin2018BERTPO}. 

Prior work on defending against rogue attacks as been explored in QA in recent years \cite{jia2017adversarial,wang2018robust,Maharana2020AdversarialAP}. The earlier work \cite{jia2017adversarial,wang2018robust} trained a BiDAF \cite{seo2016bidirectional} QA system using adversarial mutated data and most recently ~\cite{Maharana2020AdversarialAP} use reinforcement learning to select the adversarial training policies to help create a robust model. Our approach is similar to \cite{jia2017adversarial,wang2018robust}, but we also include data-augmentation of translated QA pairs in our model which helps build multi-lingual models that are less susceptible to attacks. We leave exploring more sophisticated defense strategies as future work. 

The most recent adversarial work, ~\cite{Maharana2020AdversarialAP} extends  \cite{wang2018robust} and expands the pool of distractor statements by synonyms replacement and paraphrasing. They use the distractors to attack an English QA system and they show that training with such distractors helps the system overcome these attacks. However, training with adversarial statements causes a loss in performance of the English system when evaluated on the original data, without the attack. A more sophisticated approach is needed to preserve the performance on the original dataset. We find that training our model with translated multilingual data preserves the performance of the original systems when the attack is not present.
Finally, some cross-lingual experiments with $de$, $ru$ and $tr$ are shown, but they do not use multilingual models or any multilingual attacks. They apply the translate-test method to translate to English, predict the results on the translation and align the result back to the original language.
\section{Attack Methodology}


\label{sec:methods}
\begin{figure*}[t]
\centering
    \includegraphics[width=\textwidth]{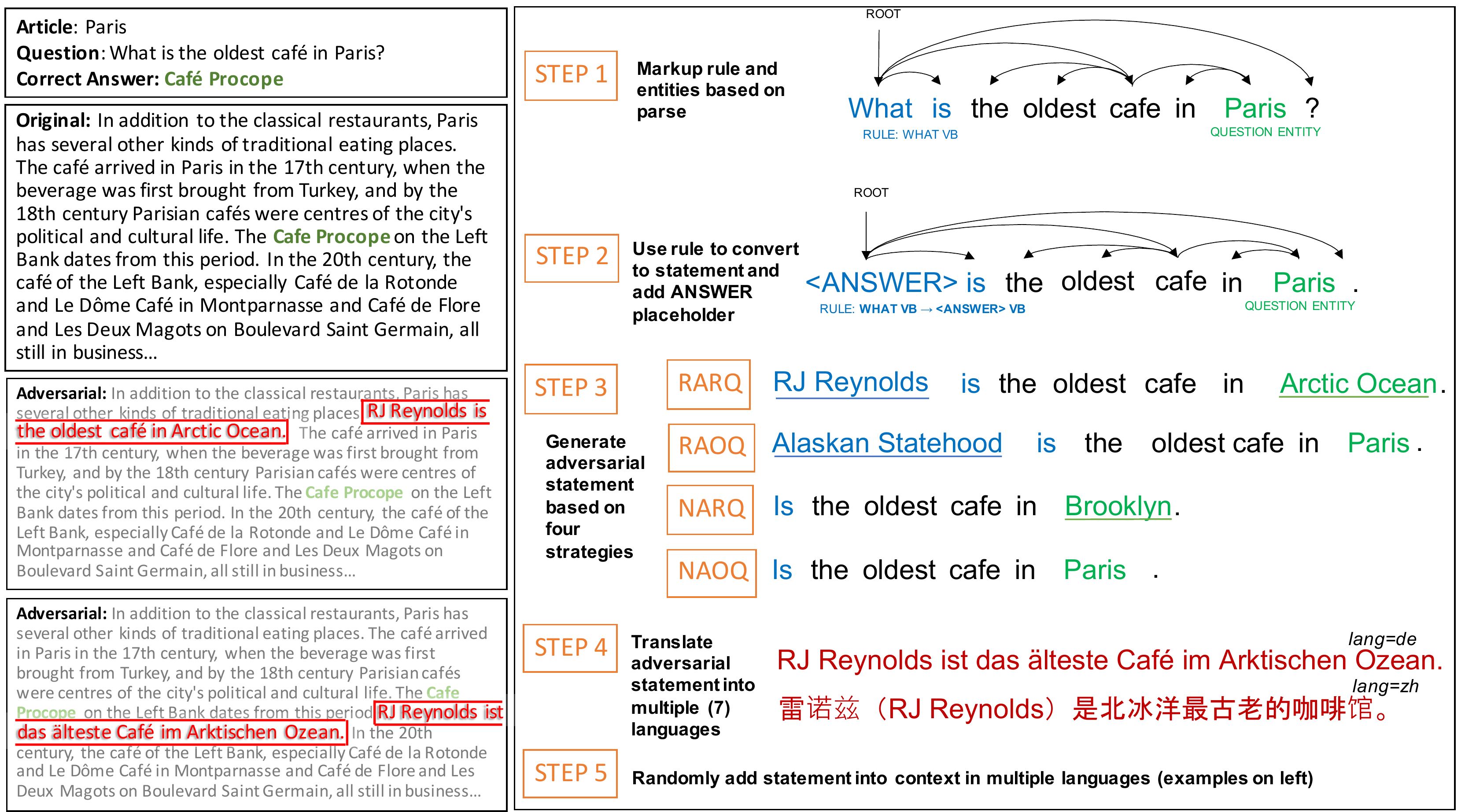}
\caption{Attack methodology example showing how to use the parse to \textcolor{blue}{generate the rule} and convert the question into a statement with \textcolor{indiagreen}{candidate question entities} to change for each of the attacks. The new random entities are \underline{underlined}. \textcolor{red}{Adversarial} statements are created in multiple languages and randomly added to the passage (on left). 
}
    \label{fig:addmlsent}
\end{figure*}




Prior QA attack strategies include adding adversarial sentences to distract RC systems \cite{jia2017adversarial,wang2018robust,wallace2019universal} and using question variations \cite{sen-saffari-2020-models}.
Our attack approach\footnote{Our adversarial code, data and pre-trained MLQA models will be released upon acceptance.} extends the AddSentDiverse algorithm \cite{jia2017adversarial,wang2018robust}) which builds adversarial distractor examples by converting the question $Q$ into a statement $S$ using several manually defined rules. 
Similarly, our goal is to generate an adversarial $S$ that is semantically similar to $Q$ but identifiable by a human reader as incorrect. We aim for grammatically correct sentences, using our generated rules but do not enforce this\footnote{AddSentDiverse fixed grammatical errors via crowd-workers to ensure the sentences are natural looking. 
We find that adversarial statements that are not fixed can also successfully attack the system, so we exclude this time-consuming and costly process.}.
We generate four different attack approaches. The attacks create various types of adversarial statements to confuse the QA system.
Further, we explore generating adversarial statements in a multi-lingual setting. The steps for converting an English $Q$ to an adversarial $S$ are shown in Figure \ref{fig:addmlsent} and described in detail below.
\subsection{Step 1: Markup Question} 

We take as input $Q$, and run two linguistic pre-processing steps: 1) Universal Dependency Parsing (UDP) \cite{mcdonald2013universal} using a model similar to \cite{qi2018universal} and 2) Named Entity Recognition (NER) \cite{sang2003introduction} using the publicly available Spacy toolkit\footnote{https://spacy.io}. We find the focus words (\eg{} which, what etc.) using their corresponding POS tags (\eg{} WRB, WB) generated by the parser. We perform a depth-first search on the parse and mark all POS tokens that are on the same level or a child of the focus word as part of the question rule. This approach creates over 9000 patterns in the SQuAD training set, some occurring only once. Frequent patterns include \textit{``what nn'', ``what vb'', ``who vb'', ``how many''}, and \textit{``what vb vb''} accounting for over 40\% of the training set. In addition, we also mark up all the entities in the question. We give priority to entities tagged by the NER that are not part of the question pattern. However, when such entities are not found, we look at nouns and then verbs to ensure better coverage. ``what vb'' is the pattern found in ``What is the oldest cafe in Paris?" as shown in Figure \ref{fig:addmlsent}.

\subsection{Step 2: Convert Question to Statement} 
The pattern found in step 1 is used to choose from 8 rules based on the common question words: \{\textit{``who'', ``what'', ``when'', ``why'', ``which'', ``where'', ``how''}\} and a catchall for any pattern that does not have question words (these are usually due to ill-formed questions or misspellings such as \textit{``Beyonce's grandma's name was?''}). The rule converts the question $Q_x$ into a statement $S_x$ containing the tagged question entities and adds an $<$ANSWER$>$ placeholder. If the first question word found in the pattern is ``what'', the rule ``what vb'' will replace ``what'' with $<$ANSWER$>$. For example, \textit{``$<$ANSWER$>$ is the oldest cafe in Paris''}. as shown in Figure~\ref{fig:addmlsent}. Sometimes, the answer is added to the end of the statement. The ``when vb vb'' pattern will trigger the rule for ``when'' which converts \textit{``When did Destiny's Child release their second album?''} to \textit{``Destiny 's Child released their second album in $<$ANSWER$>$''}.

\subsection{Step 3: Generate Adversarial Statements} 
Given question $Q$ and statement $S$, four attack statements are generated which replace \textit{$<$ANSWER$>$} and/or question entities based on the attack. The candidate entities are randomly chosen from the entities found in the SQuAD training data based on their type. The type of the answer entity is chosen based on the entity that the system predicts for the dev/test question in a non-adversarial setting. Date and number entities are not chosen from SQuAD but just randomly generated. The candidate entities are applied to create the adversarial statement using the following transformations from most complex to most simple. An example for each of the attacks is shown in Figure~\ref{fig:addmlsent}.
\\

\begin{description}
\item[RARQ: Random Answer Random Question] 
The adversarial statement has a \underline{R}andom \underline{A}nswer entity and one \underline{R}andom \underline{Q}uestion entity is changed. This approach is similar to the technique found in prior work~\cite{wang2018multi}, however they only explore monolingual. 
    
\item[RAOQ: Random Answer Original Question] 
The adversarial statement has a \underline{R}andom \underline{A}nswer entity but the \underline{O}riginal \underline{Q}uestion entities remain. This may appear to be too harsh of an adversary because the statement can look like a correct answer. However, we found that in most cases (19/20 times) people can successfully determine that this statement is adversarial 
    and it should be clear that the adversarial statement does not belong when it is in a different language than the rest of the context. We provide further discussion on this attack in Section~\ref{sec:analysis}.
    
\item[NARQ: No Answer Random Question] 
The adversarial statement has \underline{N}o \underline{A}nswer and one \underline{R}andom \underline{Q}uestion entity is changed. 

\item[NAOQ: No Answer Original Question] 
The generated adversary has \underline{N}o \underline{A}nswer and the \underline{O}riginal \underline{Q}uestion entities remain. 
\end{description}

\subsection{Step 4: Translate} 

The adversarial statements created in step 3 are always generated for English questions. Since we are evaluating a multilingual dataset and model, all statements are translated into the six other languages available in the dataset.

\subsection{Step 5: Insert Statements in Context} 

The generated statements are inserted in \textit{random} positions in $C_x$ as in \cite{wang2018multi}. This produces a new instance ($Q_x, C_y, A_y, S_z$) where $x,y,z \in L$ are the languages for the question, context, and statement respectively and they need not be the same. We avoid always adding distractors at the beginning or end of the context \cite{jia2017adversarial} as QA systems can easily learn to predict such distractors. \\
\section{Data}
\label{sec:data}





\noindent{\textbf{SQuAD v1.1}} Our primary training dataset is SQuAD v1.1 \cite{Rajpurkar_2016} which is an English only dataset. Training on English data and evaluating on a multilingual dataset is considered a zero shot (ZS) setting. The SQuAD training set is large consisting of over 87,000 QA pairs. Each instance has a question $Q$, context $C$, and an answer $A$ within the context in English: ($Q_{en},C_{en},A_{en}$).

\noindent\textbf{MT-SQuAD: Augmented Translation Data}
Since we only have the SQuAD English examples to train our system, we use a commercial translation API\footnote{https://www.ibm.com/watson/services/language-translator/} to expand our training data. For every example in SQuAD, we translate the question and the context to six other languages:  German $(de)$, Spanish $(es)$, Arabic $(ar)$, Hindi $(hi)$, Vietnamese $(vi)$ and Chinese $(zh)$. We obtain the gold answer for the translated examples by aligning the gold answer in the English context to the translated context. We perform the answer alignment by placing pseudo HTML tags around $A_{en}$ and then translate $C_{en}$. In the majority of cases, the translated answer is marked by the same tags inside the translated context. We discard the translation when the answer alignment does not succeed\footnote{Hindi is the least successful, only translating 24\% of the data. The rest translate at least 90\% of the data.}.

\noindent\textbf{MLQA Dataset}
We evaluate our model on MLQA \cite{lewis2019mlqa}, a large parallel multilingual QA dataset consisting of 7 languages.
Each MLQA instance has ($Q_{x},C_{y},A_{y}$) where the question and context language need not be the same. 
We focus on the more comprehensive task, 
 generalized cross-lingual transfer (G-XLT), 
 by extracting answer $A_y$ from context $C_y$ in language $y$ given question $Q_x$ in language $x$, for all language pairs. There is no training data in MLQA, encouraging ZS approaches using datasets such as SQuAD. 
We use the official evaluation metric of the MLQA dataset, 
the mean F1 scores 
for all $(x,y)$ language pairs. 

Note that although we focus on the SQuAD-like MLQA corpus in our experiments, our approach can be applied to any reading comprehension task and any corpus.

\section{Experimental Setup}

Our experiments focus on the MLQA task using the \mbert{} pre-trained language model. In all cases our model is the same but the training and test data used differs for each attack and defense. We train our QA models using mBERT with our two datasets described in Section~\ref{sec:data}: SQuAD ($en$ only) and MT-SQuAD. This results in two models: \bertqa{} and \mtbertqa{}.

The \bertqa{} and \mtbertqa{} model architecture is a standard MRC/QA model built on top of \mbert{} \cite{Devlin2018BERTPO} which processes two separate input sequences, one for the question and one for the given context and train two classification heads on top of \mbert{}, pointing to the start and end positions of the answer span.

We compute G-XLT at test time by extracting answer $A_z$ from context $C_z$ in language $z$ given question $Q_y$ in language $y$ for all language pairs. The test results are computed on the MLQA dataset. We use the official evaluation metric used by the MLQA dataset and report the performance of our models on the G-XLT task with the mean F1 scores for all $(Q_l,C_l)$ language pairs.

\subsection{Hyper-parameters}
We perform our hyper-parameter selection on the MLQA dev split. We use $3e-5$ as our learning rate, $384$ as maximum sequence length, and a doc stride of $128$ for all models. All other parameters are default \bertbase{} parameters \cite{Devlin2018BERTPO}. We train our models for 2 epochs for English only training, but a single epoch gave the best performance on dev for data augmentation with translation



\begin{table}[t]
    \centering
    \setlength{\tabcolsep}{2.5pt}
\resizebox{\columnwidth}{!}{\begin{tabular}{c|c|c||c|c||c|c}
    & \multicolumn{2}{c||}{SQuAD Dev} & \multicolumn{2}{c||}{MLQA Dev} & \multicolumn{2}{c}{MLQA Test} \\
	 	&	\bertbase{}	&	\mbert{} &	\bertbase{}	& \mbert{} &	\bertbase{}	&	\mbert{} 	\\
	\toprule
ORIG	&	86.0	&	89.0	&	69.2	&	80.3	&	69.2	&	80.4	\\
\midrule
RARQ	&	47.8	&	38.3	&	36.7	&	38.0	&	37.3	&	37.9	\\
NARQ	&	76.9	&	62.5	&	59.1	&	55.4	&	58.8	&	56.2	\\
RAOQ	&	\textbf{39.3}	&	\textbf{13.0}	&	\textbf{30.0}	&	\textbf{12.3}	&	\textbf{29.2}	&	\textbf{11.8}
\\ 
NAOQ	&	87.2	&	67.8	&	70.2	&	59.9	&	69.6	&	60.0	\\
\end{tabular}}
    \caption{F1 scores for the original dataset and attacking $(C_{en},Q_{en})$ with adversarial $S_{en}$ for SQuAD and MLQA using \bertbaseqa{}  and \bertqa{} models. The  attacks  that  have  the  biggest  impact  are highlighted in \textbf{bold}.}
    \vspace{-1em}
    \label{tab:en-attack}
\end{table}


\section{Attack Experiments}
\label{sec:experiments}
\subsection{English Only}
 We first explore the robustness of \bertqa{}, when trained on SQuAD, with the attacks using English data from SQUAD and MLQA to create the adversarial statement $S_{en}$ as $(Q_{en}, C_{en}, A_{en}, S_{en})$. We contrast the multilingual \bertqa{} to the English BERT-Base (\bertbaseqa{}) 
on the same attacks. 
Table~\ref{tab:en-attack} shows the results for these experiments. We find that the attack has a significantly bigger impact when using the \bertqa{} model than using \bertbaseqa{} for the SQuAD dataset, even though \bertqa{} performs 3 points better than \bertbaseqa{} on the original (ORIG) dataset that does not contain adversarial statements. Similarly, the attack with the MLQA dataset is far more successful on the \bertqa{} model compared to \bertbaseqa{}. Interestingly, the NAOQ attack is not successful on both datasets when using \bertbaseqa{}. We suspect that the model understands that no answer entity is available and the syntax of the statement helps steer the model to the correct answer, but the same cannot be said when using the \bertqa{} model. 
Most attacks on the \bertqa{} model with MLQA are less effective than the the attacks with the SQuAD dataset, which may be explained by the differences between the 
the MLQA test data and the SQuAD training data.

\begin{table}[t]
    \centering
    \small
    \begin{tabular}{l|r|r|r|r|r}
	 & $S_{C_l}$ &	 $S_{Q_l}$	&	$S_{en}$	&	$S_{de}$	&	$S_{zh}$	\\
	 \toprule
 	&	\multicolumn{5}{c}{\bertqa{}} \\
 	\hline
 	\midrule
ORIG	&	52.1	&	52.1	&	52.1	&	52.1	 &	52.1	\\
\midrule
RARQ	&	36.3		&	15.7	&	27.7	&	33.1	&	37.8\\
NARQ	&	41.0	&	24.8		&	36.8	&	38.8	&	41.5\\
RAOQ	&	\textbf{24.8}	&	\textbf{7.5}		&	\textbf{17.2}	&	\textbf{23.2}	&	\textbf{29.5}\\
NAOQ	&	39.5 	&	26.8		&	36.8	&	36.7	&	38.7\\
\toprule
& \multicolumn{5}{c}{\mtbertqa{}}		\\
\hline
\midrule
ORIG	&	58.3	&	58.3	&	58.3	&	58.3	&	58.3	\\
\midrule
RARQ	&	36.6	&	18.8	&	27.7	&	39.0	&	20.3	\\
NARQ	&	42.7	&	30.3	&	36.6	&	43.8	&	29.1	\\
RAOQ	&	\textbf{22.7}	&	\textbf{9.2}	&	\textbf{16.1}	&	\textbf{28.1}	&	\textbf{9.3}	\\
NAOQ	&	41.2	&	30.1	&	33.7	&	40.3	&	27.8	\\
\bottomrule
\end{tabular}
    \caption{G-XLT results for the original test set and when attacking the system using each attack with the \bertqa{} and \mtbertqa{} models with statements in different languages. The column headers refer to the language $x$ of the adversarial statement as $S_x$. The attacks that have the biggest impact are highlighted in \textbf{bold}.}
    \label{tab:mlqa-s_y-attack}
    \vspace{-1em}
\end{table}

\subsection{Multilingual} Our main focus in this paper is to show the impact of attacking the system in a multilingual setting. We are using the MLQA dataset to evaluate the attacks  and show all multilingual results in Table \ref{tab:mlqa-s_y-attack}. We attack both \bertqa{}, a multilingual system trained with English only, and also \mtbertqa{}, a multilingual system trained with data in 6 languages, and we show that both systems are affected by the attacks. We explore generating our four types of attacks using statements based on five language setups: The language of the context $S_{C_l}$, English $S_{en}$, German $S_{de}$, Chinese $S_{zh}$, and the language of the question $S_{Q_l}$. 

First, we explore the natural setting of adding an adversarial statement in the language of the context, $S_{C_l}$, as ($Q_x,C_y,A_y,S_y)$ where $x,y \in L$ and may or may not be the same language. 
This attack reduces performance of both \bertqa{} and \mtbertqa{} by at least 11 points. The model has an easier time distinguishing what the correct answer is when the adversarial statement is in the same language as the context.  

Next, we look at adding the adversarial statement in the language of the question, $S_{Q_l}$ as ($Q_x,C_y,A_y,S_x)$ where $x,y \in L$ and may or may not be the same language. 
These attacks have the strongest impact, indicating that the model gives a preference to the language in the question regardless of whether it matches the language in the context.

We also explore adding an English adversarial statement, $S_{en}$, to the context $C_y$ for \textit{all} of the test examples as $(Q_x,C_y,A_y,S_{en})$.  English is the language of training data in \bertqa{}.  
Individual per language results (not visible in the table) showed that 
it causes a significant decrease in performance for all attacks ranging from 13.3 to 37 points compared to the original test set. 
Furthermore, adding $S_{en}$ to $C_y$ where $y \neq en$ also causes a significant decrease in performance similar to or worse than $C_{en}$, even though the statement is in a different language than the rest of the context and the question may or may not be in English. 

We explore this further using machine translation to add the adversarial statement in German, $S_{de}$ and Chinese, $S_{zh}$. 
The $S_{de}$ attack is slightly weaker than the $S_{en}$ attack and the $S_{zh}$ attack is considerably weaker than the $S_{en}$ attack; the model has a strong preference for predicting English statements regardless of the language of $Q_x$ and $C_y$ and prefers languages more similar to English than those that are more distant. 
Furthermore, the preference for English compared to German is also observed when using \mtbertqa{} which is trained with data in 6 languages. However, the attack on Chinese is much stronger when using \mtbertqa{} compared to \bertqa{} indicating that the model does learn to distinguish between more distant languages in a non-ZS approach. 
We also notice that with \mtbertqa{} the non-adversarial model performs significantly better, but the attack is stronger. This may be due to the added noise from translation.
\subsection{Attack Strategies} Our four different strategies for generating adversaries  perform similarly across all experiments. RAOQ is consistently the strongest attack because it looks the most like the actual answer even though humans can easily identify that it is not the correct answer. At minimum, it causes a 30 point reduction (See RAOQ $S_{de}$ attack in \mtbertqa{} model in Table~\ref{tab:mlqa-s_y-attack}) but in its worst attack it causes a whopping \textit{49 point} reduction in F-score (See RAOQ $S_{zh}$ attack in \mtbertqa{} model in Table~\ref{tab:mlqa-s_y-attack}). RAOQ has a significantly stronger impact than RARQ which is similar to prior work \cite{wang2018robust}. Still, RARQ is the second strongest attack even though the answer entity and a question entity are randomly changed. NAOQ and NARQ are the weakest because no answer is in the statement. These appear less grammatical and it is easier for the model to avoid them. Still, even these attacks are very successful, causing at minimum a 10 point reduction in F-score as shown in Table~\ref{tab:mlqa-s_y-attack}.

\section{Defense}

We have shown that we can successfully attack the system using several scenarios and exploit vulnerabilities in multilingual QA models. Now we follow prior approaches\cite{jia2017adversarial,wang2018robust,Maharana2020AdversarialAP}, to explore how we can make the model more robust to multilingual adversarial attacks. We achieve this by augmenting the model with examples of the attack strategies during training in addition to the original (ORIG) data so that it can learn to spot these adversaries on its own,  as described in \cite{Maharana2020AdversarialAP}. To avoid the over-fitting of the statement structure we add randomness to the template and we also randomize the location of the points of confusion inside the context. 

Our approach for generating statements is the same as during the attack strategy, except that in contrast to the attack strategy, in this case we choose random entities to replace the answer using the gold answer instead of the prediction. We train a separate model for each attack ORIG + \{RARQ, NARQ, RAOQ, NAOQ\}, and then evaluate how the new models trained with the adversarial data perform on the various adversarial test sets.  We explore the defense with both the \bertqa{} and \mtbertqa{} models. In the \bertqa{} model all attack statements during training time are in English. In the \mtbertqa{} model they are in all languages.
The results are shown in Table~\ref{tab:defense}. The ORIG column shows the original training model with the attacks as in Table~\ref{tab:mlqa-s_y-attack} for comparison. 

The ORIG rows of Table~\ref{tab:defense} show how the  models perform when there are no adversarial statements. The defense models that contain the adversarial statements suffer a loss (2-4 points) in performance when testing with \bertqa{} on the original data. On the other hand, all \mtbertqa{} defense models obtain consistent scores and maintain the performance of the original model when testing with the original dataset where no adversarial statements exist.  We also find that all the defense training models can effectively protect the QA system against the attacks, in most cases reaching near original performance if the attack is the same as defense strategy (the diagonal results; \eg{} the RARQ training model recovers best for the RARQ test set), with consistently better performance for the \mtbertqa{} model. The models can even recover well for different attack strategies. This indicates that the defense models may be able to withstand other types of attacks than the ones tried here. For the \bertqa{} model RAOQ recovers better using all other training models. We expect the similarity to the question causes difficulty during training. 

Initial experiments applying all attacks at once was slightly better than the individual models indicating that such an approach could be useful. Our insights show that basic defense strategies assist with over-fitting of multilingual QA models that are under attack. We show particularly positive success when augmenting the MT model with attack statements during training. We leave exploring more complex models such as adversarial networks \cite{NIPS2014_5423,goodfellow2015explaining} on top of BERT as future work.

\begin{table}[t]
    \centering
    \small
    \setlength{\tabcolsep}{2pt}
    \resizebox{\columnwidth}{!}{
    \begin{tabular}{c|c||r|r|r|r|r}
    Test & \textit{S} & \multicolumn{5}{c}{Training Data} \\
	Data	& Lang	&	ORIG	&	+RARQ	&	+NARQ	&	+RAOQ	&	+NAOQ	\\
\toprule
 \multicolumn{7}{c}{\mbert{}: SQuAD Training Data} \\
 \hline
\midrule ORIG	&	&	52.1	&	50.1	&	49.6	&	48.8	&	48.3	\\
\midrule	&	$S_{en}$	&	27.7	&	46.9	&	42.4	&	42.3	&	32.1	\\
 RARQ	&	$S_{Q_l}$	&	15.7	&	44.5	&	38.0	&	35.7	&	21.7	\\
        &   $S_{C_l}$  & 36.3	 & 44.0   &	40.7  &	40.0  &	35.5 \\
\midrule	&	$S_{en}$	&	36.8	&	\textbf{47.1}	&	\textbf{48.2}	&	\textbf{45.1}	&	\textbf{48.8}	\\
 NARQ	&	$S_{Q_l}$	&	24.8	&	44.4	&	45.9	&	40.2	&	35.3	\\
        &   $S_{C_l}$ &  41.0  &	44.6  &	45.0  &	42.2  &	40.0 \\
\midrule	&	$S_{en}$	&	17.2	&	36.1	&	34.1	&	38.0	&	25.5	\\
 RAOQ	&	$S_{Q_l}$	&	\textbf{7.5}	&	33.3	&	30.9	&	33.9	&	17.9	\\
        &   $S_{C_l}$   & 24.8   &	32.7 &	31.4  &	32.9  &	27.3 \\
\midrule	&	$S_{en}$	&	36.8	&	42.3	&	46.9	&	43.6	&	44.0	\\
 NAOQ	&	$S_{Q_l}$	&	26.8	&	38.1	&	45.3	&	41.1	&	42.4	\\
        &  $S_{Q_c}$ & 39.5 &	40.0 &	42.5  &	39.9   & 40.4 \\
\hline
\toprule
\multicolumn{7}{c}{\mtbertqa{}: MT Training Data} \\
\hline
\midrule ORIG	&	&	58.3	&	57.9	&	58.0	&	58.0	&	58.0	\\
\midrule\textbf{}	&	$S_{en}$	&	18.8	&	56.8	&	51.0	&	55.4	&	33.1	\\
 RARQ	&	$S_{Q_l}$	&	20.3	&	\textbf{57.6}	&	54.0	&	\textbf{57.1}	&	37.2	\\
 &   $S_{C_l}$ &   36.6	&   57.1	&   53.3 &	56.4	& 43.5 \\
\midrule	&	$S_{en}$	&	30.3	&	56.1	&	\textbf{57.6}	&	56.5	&	54.3	\\
 NARQ	&	$S_{Q_l}$	&	29.1	&	56.7	&	57.5	&	57.0	&	54.8	\\
 & $S_{C_l}$ & 42.7 &	56.3 &	57.3 & 56.6	 & 54.6 \\
\midrule	&	$S_{en}$	&	\textbf{9.2}	&	47.2	&	40.9	&	54.1	&	29.5	\\
 RAOQ	&	$S_{Q_l}$	&	9.3	&	52.1	&	47.1	&	56.9	&	35.5	\\
        &  $S_{C_l}$ & 22.7 &	50.3 &	45.2 &	55.1 &	38.0 \\
\midrule	&	$S_{en}$	&	30.1	&	48.7	&	56.1	&	55.2	&	\textbf{58.0}	\\
 NAOQ	&	$S_{Q_l}$	&	27.8	&	51.1	&	56.6	&	56.5   &	57.8	\\
        &   $S_{C_l}$  &  41.2  &	50.8  &	56.0  &	55.9  &  57.8 \\
	\bottomrule
    \end{tabular}}
    \caption{Average G-XLT results for the Defense models for the SQuAD and MT training data per test set and the statement language \textit{S}. The first column (ORIG) shows the attacks as found in Table~\ref{tab:mlqa-s_y-attack} for comparison. The header row shows the different training strategies which include the original data and one attack strategy. The attack strategy during test time is shown by row as strategy + $S$ Lang. The defense models per column with the biggest improvement are highlighted in \textbf{bold}.}
    \label{tab:defense}
    \vspace{-1em}
\end{table}


\section{Discussion and Analysis}
\label{sec:analysis}

We have shown that our attacks that are generated automatically by converting the question to a statement are successful in fooling the system. It is important that the attacks are easily detectable by people to ensure that our attacks are fair in exploiting flaws in the system. Specifically, with the RAOQ attack, these can appear to be natural sentences that could answer the question. If outside knowledge is unknown, the actual answer and the adversarial RAOQ statement can both appear correct without context. However, these sentences will usually not flow with the rest of the passage and it will be the only mention of the entity. The correct entity will likely be mentioned more than once in the passage. Finally, to the average person with external knowledge it should be easy to recognize that the entities do not belong with the rest of the passage. We verified this through an annotation task on 100 English instances using two native English speakers (50 each), 20 for each of the four attacks and 20 for the original instances with no attacks. The annotators were first asked to just look at the passage without seeing the question and say which sentence did not belong. The annotators detected the adversarial statement or lack of 95\% of the time. Next, we allowed them to see the question (information the model has as well), and the annotators were able to detect the adversarial statement or lack of 99\% of the time indicating that people can easily detect the adversarial statement. 

We provide the following example of a passage with the RAOQ English statement in an English passage to illustrate that it is usually clear to the average English speaker which sentence is the adversarial statement:

\noindent\textbf{Question:} Which artist created the mural on the Theodore M. Hesburgh Library?

\textbf{Context:} The library system of the university is divided between the main library and each of the colleges and schools.\textcolor{red}{\textit{ The National Declassification Center artist created the mural on the Theodore M. Hesburgh Library}}. The main building is the 14-story Theodore M. Hesburgh Library, completed in 1963, which is the third building to house the main collection of books. The front of the library is adorned with the Word of Life mural designed by artist Millard Sheets. This mural is popularly known as \"Touchdown Jesus\" because of its proximity to Notre Dame Stadium and Jesus' arms appearing to make the signal for a touchdown.

In the above passage the adversarial statement is in \textit{\textcolor{red}{italics}}. First, the entity \textit{National Declassification Center} that is included in the sentence is clearly not the name of the artist. Further, even if one was uncertain about this, the sentence doesn't flow with the rest of the paragraph. It reads much better if that sentence is removed. Finally, the sentence that does include the correct answer flows better on its own, and within the paragraph.

\section{Conclusion}

We have shown several novel adversaries that successfully attack \mbert{} for MLQA. Specifically, we show that the language of the adversarial statement impacts the attack with priority given to English and the language of the question regardless of the other languages in the QA pair. We also show that including such attack strategies while training our defense brings back performance without the need for complex neural network engineering. Not only do the strategies improve results for their corresponding attack, they help for all our attacks indicating model robustness. In the future, we plan to expose vulnerabilities on other multilingual LMs and datasets and explore more sophisticated defense strategies.


\section{Ethics}

The intended use of this work is to expose vulnerabilities in question answering systems solely as a means of making them more robust. We are aware that this work could be misused to attack question answering systems. We do not promote attacking systems for any malicious intent.

\bibliography{main}
\bibliographystyle{acl_natbib}

\end{document}